\documentclass[sigconf]{acmart}

\AtBeginDocument{%
  \providecommand\BibTeX{{%
    \normalfont B\kern-0.5em{\scshape i\kern-0.25em b}\kern-0.8em\TeX}}}
\usepackage{multirow}
\usepackage{multicol}
\usepackage{xspace}
\usepackage{bbm}
\usepackage{colortbl}
\usepackage{amsmath}
\newcommand{\model}{{ExcelFormer}\xspace}
\newcommand{\featmix}{\textsc{Feat-Mix}\xspace}
\newcommand{\hidmix}{\textsc{Hid-Mix}\xspace}
\definecolor{camel}{rgb}{0.94, 0.82, 0.62}
\definecolor{lightgray}{rgb}{0.9, 0.9, 0.9}

\copyrightyear{2024}
\acmYear{2024}
\setcopyright{acmlicensed}\acmConference[KDD '24]{Proceedings of the 30th ACM SIGKDD Conference on Knowledge Discovery and Data Mining}{August 25--29, 2024}{Barcelona, Spain}
\acmBooktitle{Proceedings of the 30th ACM SIGKDD Conference on Knowledge Discovery and Data Mining (KDD '24), August 25--29, 2024, Barcelona, Spain}
\acmDOI{10.1145/3637528. 3671893}
\acmISBN{979-8-4007-0490-1/24/ 08}

\begin{document}

\title{\model: A neural network surpassing GBDTs on tabular data}
\author{Jintai Chen$^*$}
\email{jtchen721@gmail.com}
\affiliation{%
  \institution{\institution{Univ. of Illinois Urbana-Champaign}}
  \country{Urbana, IL, USA}
}

\author{Jiahuan Yan$^*$}
\thanks{*: Co-first Authors.}
\email{jyansir@zju.edu.cn}
\affiliation{%
  \institution{Zhejiang University}
  \country{Hangzhou, China}
}

\author{Qiyuan Chen}
\email{chenqiyuan1012@gmail.com}
\affiliation{%
  \institution{Zhejiang University}
  \country{Hangzhou, China}
}

\author{Danny Z. Chen}
\email{dchen@nd.edu}
\affiliation{%
 \institution{University of Notre Dame}
  \country{South Bend, IN, USA}
}

\author{Jian Wu}
\email{wujian2000@zju.edu.cn}
\affiliation{%
  \institution{Zhejiang University}
  \country{Hangzhou, China}
}

\author{Jimeng Sun}
\email{jimeng@illinois.edu}
\affiliation{
  \institution{Univ. of Illinois Urbana-Champaign}
  \country{Urbana, IL, USA}
}

\renewcommand{\shortauthors}{Jintai Chen and Jiahuan Yan, et al.}

\begin{abstract}
Data organized in tabular format is ubiquitous in real-world applications, and users often craft tables with biased feature definitions and flexibly set prediction targets of their interests. Thus, a rapid development of a robust, effective, dataset-versatile, user-friendly tabular prediction approach is highly desired. While Gradient Boosting Decision Trees (GBDTs) and existing deep neural networks (DNNs) have been extensively utilized by professional users, they present several challenges for casual users, particularly: (i) the dilemma of model selection due to their different dataset preferences, and (ii) the need for heavy hyperparameter searching, failing which their performances are deemed inadequate. 
In this paper, we delve into this question: Can we develop a deep learning model that serves as a \textit{sure bet} solution for a wide range of tabular prediction tasks, while also being user-friendly for casual users? We delve into three key drawbacks of deep tabular models, encompassing: (P1) lack of rotational variance property, (P2) large data demand, and (P3) over-smooth solution. We propose \model, addressing these challenges through a \textit{semi-permeable} attention module that effectively constrains the influence of less informative features to break the DNNs' rotational invariance property (for P1), data augmentation approaches tailored for tabular data (for P2), and attentive feedforward network to boost the model fitting capability (for P3). These designs collectively make \model a \textit{sure bet} solution for diverse tabular datasets. Extensive and stratified experiments conducted on real-world datasets demonstrate that our model outperforms previous approaches across diverse tabular data prediction tasks, and this framework can be friendly to casual users, offering ease of use without the heavy hyperparameter tuning. The codes are available at \url{https://github.com/whatashot/excelformer} and all the datasets are available at \url{https://huggingface.co/datasets/jyansir/excelformer}.
\end{abstract}




\received{8 Feb 2024}
\received[accepted]{17 May 2024}

\maketitle
\section{Introduction}\label{sec:intro}
\begin{figure*}[t]
    \centering
    \includegraphics[width=0.96\textwidth]{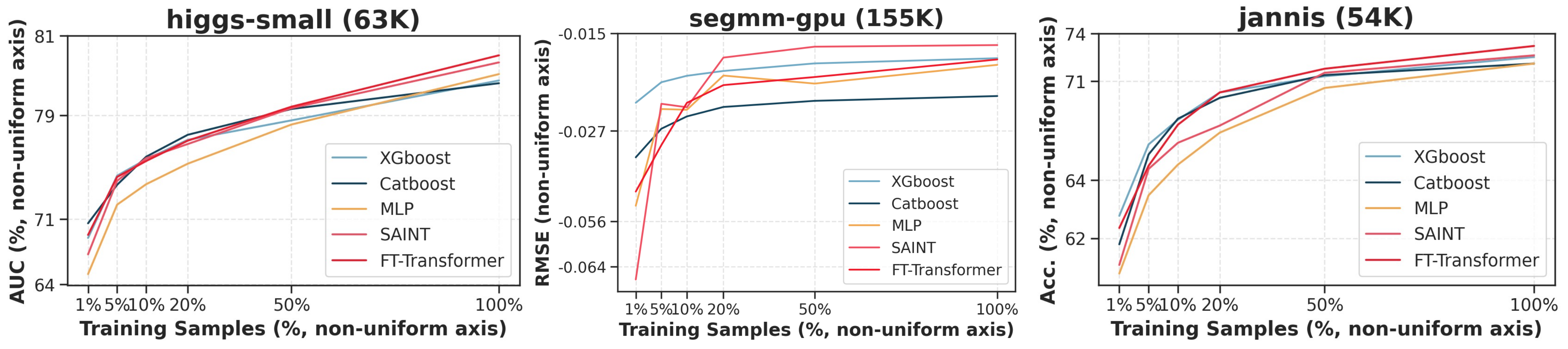}
    \vskip -1 em
    \caption{Performance variation with different percentages of training samples on three large-scale tabular datasets (total training sample count in parentheses). DNNs often exhibit better performance with larger training sample sizes, whereas GBDTs perform better when data is scarce.}\label{fig1}
    \vskip -1 em
\end{figure*}
Tabular data is ubiquitous and plays a critical role in real-world applications, spanning diverse domains, such as medical prediction~\cite{wu2022machine,yan2024making}, market prediction~\cite{wang2017deep}, and financial risk forecasting~\cite{kim2020can}.
However, unlike fields such as image and natural language processing, where data from disparate datasets frequently exhibit similar spatial or sequential feature relations and aligned semantics, tabular data often lack such ``common'' and ``standard'' data structures. 
Tables are typically created by casual users for diverse purposes. The features and targets can be defined subjectively, and table columns (features) are added or removed arbitrarily, even sometimes resulting in missing information or adding noise.
Therefore, while some bespoke frameworks following specific inductive biases have thrived in the domains of image and textual data, achieving comparable success on tabular data is notably challenging.

Therefore, users are compelled to undergo computationally intensive hyperparameter searching in model development for specific tabular datasets, and there is currently no universally recognized method for selecting a model and determining a set of hyperparameters without comprehensive testing on the target datasets. In this paper, we endeavor to design a DNN framework that serves as a \textit{sure bet} solution for diverse tabular datasets, by solving key drawbacks of existing DNNs. Inspired by \cite{grinsztajntree}, we summarize three key drawbacks of current deep tabular models, including:

\textbf{(P1) lack of rotational variance property.} 
As each table column holds distinct semantics and tabular data lack rotational invariance~\cite{ng2004feature}, rotational variant algorithms like decision trees are more efficient on tabular datasets. However, DNNs are a kind of rotationally invariant algorithms that has a worst-case sample complexity growing at least linearly in the number of uninformative features. As mentioned above, tables are created by casual users who frequently include uninformative features, underscoring the importance of rotational variance property.

\textbf{(P2) large data demand.} DNNs typically possess larger hypothesis spaces, necessitating more training data to obtain robust performances. Thus, it is widely noted that DNNs frequently exhibit competitive, and at times, superior performance compared to GBDTs on large-scale tabular datasets. Yet, their performances tend to be subpar on smaller datasets.

\textbf{(P3) over-smooth solution.} Observations suggest that DNNs tend to produce overly smooth solutions, a factor pinpointed by~\cite{grinsztajntree} as a contributor to suboptimal performance on datasets featured by irregular decision boundaries (as illustrated in Fig.~\ref{fig:pre-exp}). In contrast, decision trees partition the feature space based on thresholds along each axis,  resulting in sharp boundaries that are demonstrated to be more suitable for a majority of datasets.
\begin{figure*}[ht]
    \centering
    \vskip -0.8 em
    \includegraphics[width=0.9\textwidth]{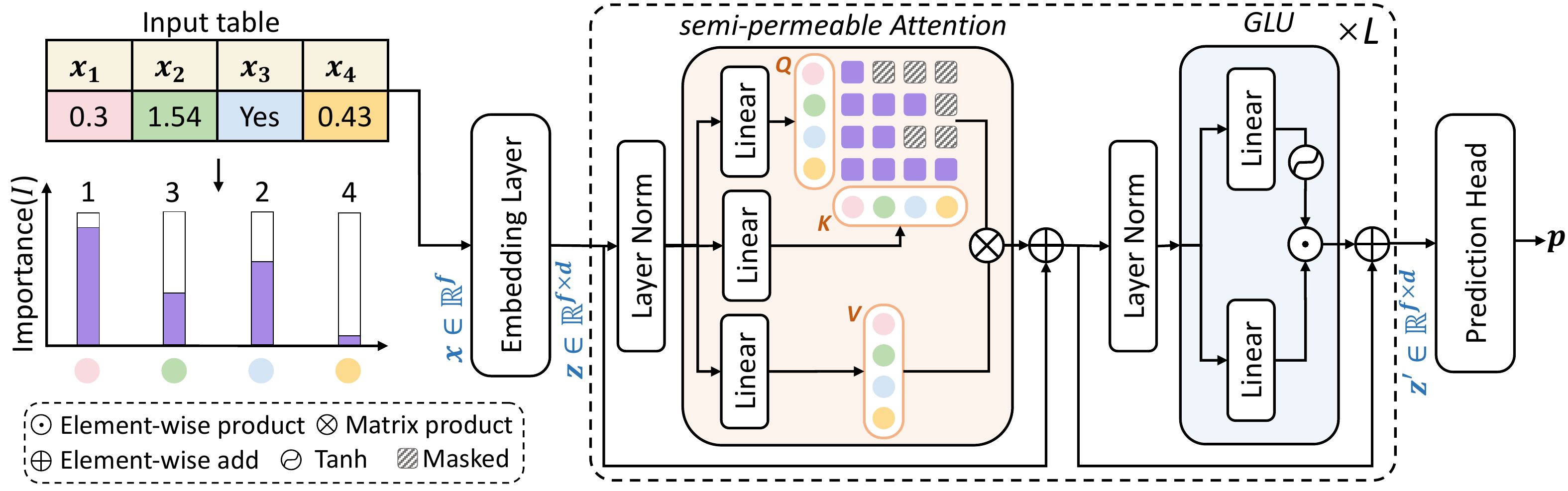}
    \vskip -0.4 em
    \caption{An illustration of our proposed \model model. Diverse feature types are pre-processed procedure as in~\cite{rubachev2022revisiting}, followed by quantile transformation and embedding layer to convert them into numerical embeddings. Each feature embedding $z_i\in \mathbb{R}^d$ serves as a token in \model.}
    \label{fig:framework}
    \vskip -1. em
\end{figure*}

Among these, the large data demand (P2) should be the primary obstacle to current deep tabular models: Intuitively, rotationally-invariant DNNs are data-inefficient~\cite{grinsztajntree}, but this drawback can be mitigated if there are sufficient training data~\cite{ng2004feature}. Moreover, modern DNNs have been demonstrated to be able to fit any functions~\cite{goodfellow2016deep}. Thus, if a plethora of discrete data points adequately fulfill the feature space to accurately represent the underlying feature-target functions, DNNs can definitely fit such functions rather than obtain overly smooth solutions. We empirically observed that even though DNNs outperform GBDTs on large tabular datasets, their performance significantly declines and falls short of GBDTs when fewer training samples are used. Three examples are presented in Fig.~\ref{fig1}. Thus, boosting the effectiveness of DNNs on small datasets is the key to achieve a \textit{sure bet} model.

In this paper, we delve into addressing the limitations of existing DNNs and present a robust model, \model, for diverse tabular data prediction tasks. To address \textbf{(P1)}, we design a novel attention module named the \textit{semi-permeable attention} (SPA), which selectively permits more informative features to gather information from less informative ones. This results in a noticeably reduced influence of less informative features. A special \textit{interaction-attenuation initialization} approach is devised to boost this module. This initialization approach sets SPA's parameters with minimal values, effectively attenuating tabular data feature interactions during the initial stages of training. Consequently, SPA undertakes more cautious feature interactions at the beginning, learning key interactions and aiding \model to break the rotational invariance. 

To address \textbf{(P2)}, we introduce two interpolation-based data augmentation approaches for tabular data: \featmix and \hidmix. Interpolation-based data augmentation approaches, such as Mixup~\cite{zhang2018mixup} and its variants~\cite{verma2019manifold,uddin2020saliencymix}, have demonstrated their effectiveness in computer vision tasks. However, as the feature-target functions are often irregular~\cite{grinsztajntree}, simple interpolation methods, like Mixup, tend to regularize DNNs to behave linearly in-between training examples~\cite{zhang2018mixup}, which potentially conflicts with the irregular functions. Therefore, they often fall short of improving and even degrading the performance of DNNs. We propose two data augmentation approaches, \featmix and \hidmix, to avoid such conflicts and respectively encourage DNNs to learn independent feature transformations and to conduct sparse feature interactions.

To address \textbf{(P3)}, we employ an attentive feedforward network to replace the vanilla multilayer perceptron-based feedforward network in the Transformer. Both consisting of two fully connected layers, this substitution integrates an attentive module to enhance the model's ability to effectively fit irregular boundaries. Following previous work, we utilize Gated Linear Units (GLU)~\citep{shazeer2020glu} and do not introduce any new modules. Experimental results confirm that this simple substitution effectively improves model performances.

Notably, replacing the vanilla self-attention and feedforward network with the SPA and the attentive feedforward network does not incur any additional computational burden, and the data augmentation approaches come at negligible cost. That means, the proposed \model maintains a comparable size to that of cutting-edge tabular Transformers (e.g., FT-Transformer~\citep{rubachev2022revisiting}). Comprehensive and stratified experiments have demonstrated the superiority of our designed \model:
\begin{enumerate}
    \item Our model outperforms existing GBDTs and DNNs not only on small tabular datasets where existing DNNs typically underperform GBDTs, but also on large-scale datasets where traditional DNNs have shown preference (Sec.~\ref{sec:resultanddiscussion}).
    \item Across a spectrum of stratified experiments in terms of feature quantity, dataset scale, and task types, our \model consistently outperforms both GBDTs and DNNs. Additionally, we observe that besides our \model, existing approaches excel in handling different types of datasets.
    This underscores its status as a reliable solution for non-expert users, mitigating the need for intricate model selection (Sec.~\ref{3232} and Sec.~\ref{3443}).
    \item Notably, while most existing approaches necessitate intensive hyperparameter tuning through repeated model runs (typically 50-100 times), our model achieves superior performance with pre-fixed parameters, offering a time-efficient and user-friendly solution (Appendix~\ref{app:time}).
\end{enumerate}
\section{Related Work}\label{sub:rw}
\subsection{Supervised Tabular Data Prediction} 
While deep neural networks (DNNs) have proven to be effective in computer vision~\cite{khan2022transformers} and natural language processing~\cite{vaswani2017attention}, GBDT approaches like XGBoost continue to be the preferred choice for tabular data prediction tasks~\cite{katzir2020net,grinsztajntree}, particularly on smaller-scale datasets, due to their consistently superior performance.
To enhance the performance of DNNs, recent studies have focused on developing sophisticated neural modules for (i) handling heterogeneous feature interactions~\cite{gorishniy2021revisiting,chen2022danets,T2G,tabcaps}, (ii) seeking for decision paths by emulating decision-tree-like approaches~\cite{katzir2020net,popov2019neural,arik2021tabnet}, or (iii) resorting to conventional approaches~\cite{cheng2016wide,guo2017deepfm} and regularizations~\citep{jeffares2023tangos}. In addition to model designs, various feature representation approaches, such as feature embedding~\cite{gorishniy2022embeddings,tabcaps}, discretization of continuous features~\cite{guo2021embedding,wang2020transparent}, and Boolean algebra based methods~\cite{wang2021scalable}, were well explored.
All these efforts suggested the potentials of DNNs, but they have not yet surpassed GBDTs in performance, especially on small-scale datasets.
Moreover, there were several attempts~\cite{wang2022transtab,arik2021tabnet,yoon2020vime,zhu2023xtab} to apply self-supervision learning on tabular datasets. However, many of these approaches are dataset- or domain-specific, and transferring these models to distant domains remains challenging due to the heterogeneity across tabular datasets. While pretrained on a substantial dataset corpus, XTab~\cite{zhu2023xtab} offered only a modest performance improvement due to the limited shared knowledge across datasets. 
TapPFN~\cite{tabpfn} concentrated on solving classification problems for small-scale tabular datasets and achieved commendable results.
However, its efficiency waned when applied to larger datasets and regression tasks. In summary, compared to decision tree-based GBDTs, DNNs still fall short on tabular data, especially on small-scale ones, which remains an open challenge.



\subsection{Mixup and other Data Augmentations}

The vanilla Mixup~\cite{zhang2018mixup} generates a new data through convex interpolations of two existing data, which was proved beneficial on computer vision tasks~\cite{tajbakhsh2020embracing,touvron2021training}. However, we have observed that vanilla Mixup may conflict with irregular target patterns (please refer to Fig.~\ref{fig:pre-exp}) and typically achieves inferior performance. For instance, in the context of predicting therapy feasibility, a 70-year-old man (elderly individual) and a 10-year-old boy (young individual) may not meet the criteria for a particular therapy, but an individual with an interpolated feature value (aged 40) would benefit from it. Namely, the vanilla Mixup can lead to over smooth solution, which is considered to be unsuitable~\cite{grinsztajntree}. ManifoldMix~\cite{verma2019manifold} applied similar interpolations in the hidden states, which did not fundamentally alter the data synthesis approach of Mixup and exhibited similar characteristics to the vanilla Mixup. The follow-up variants CutMix~\cite{yun2019cutmix}, AttentiveMix~\cite{walawalkar2020attentive}, SaliencyMix~\cite{uddin2020saliencymix}, ResizeMix~\cite{qin2020resizemix}, and PuzzleMix~\cite{kim2020puzzle} spliced image pieces spatially, preserving local image patterns but being not directly applicable to tabular data. CutMix is used in SAINT~\cite{somepalli2021saint} for tabular data prediction, but it is highly impacted by uninformative features, as shown in Table~\ref{tab:noisy}.
\citet{kadra2021well} investigated various data augmentation techniques aimed at enhancing the performance of MLPs on tabular data. However, these methods were found to be effective only on a limited number of tabular datasets, requiring time-consuming enumeration and testing of these options. In contrast, this paper introduced two novel data augmentation approaches for tabular data, \hidmix and \featmix, which avoid the conflicts encountered with Mixup and contribute to \model achieving superior performance.
\section{\model}\label{sec:model}
The workflow of \model is illustrated in Fig.~\ref{fig:framework}. \model processes data by the following components: 1) After pre-processing like~\citet{gorishniy2021revisiting}, the embedding layer featurizes and embeds tabular features to token-level embeddings; 2) token-level embeddings were alternately processed by the newly proposed \textit{semi-permeable attention} module (SPA) and gated linear units (GLUs). 3) Finally, a prediction head yields the final target. In the following, we will introduce the novel \textit{semi-permeable attention} with the interaction attenuated initialization and the GLU based attentive feedforward network first and then the rest parts of \model.
\subsection{Solving (P1) with Semi-Permeable Attention}\label{sec:diam}
\begin{figure}[t]
    \centering
    \includegraphics[width=0.45\textwidth]{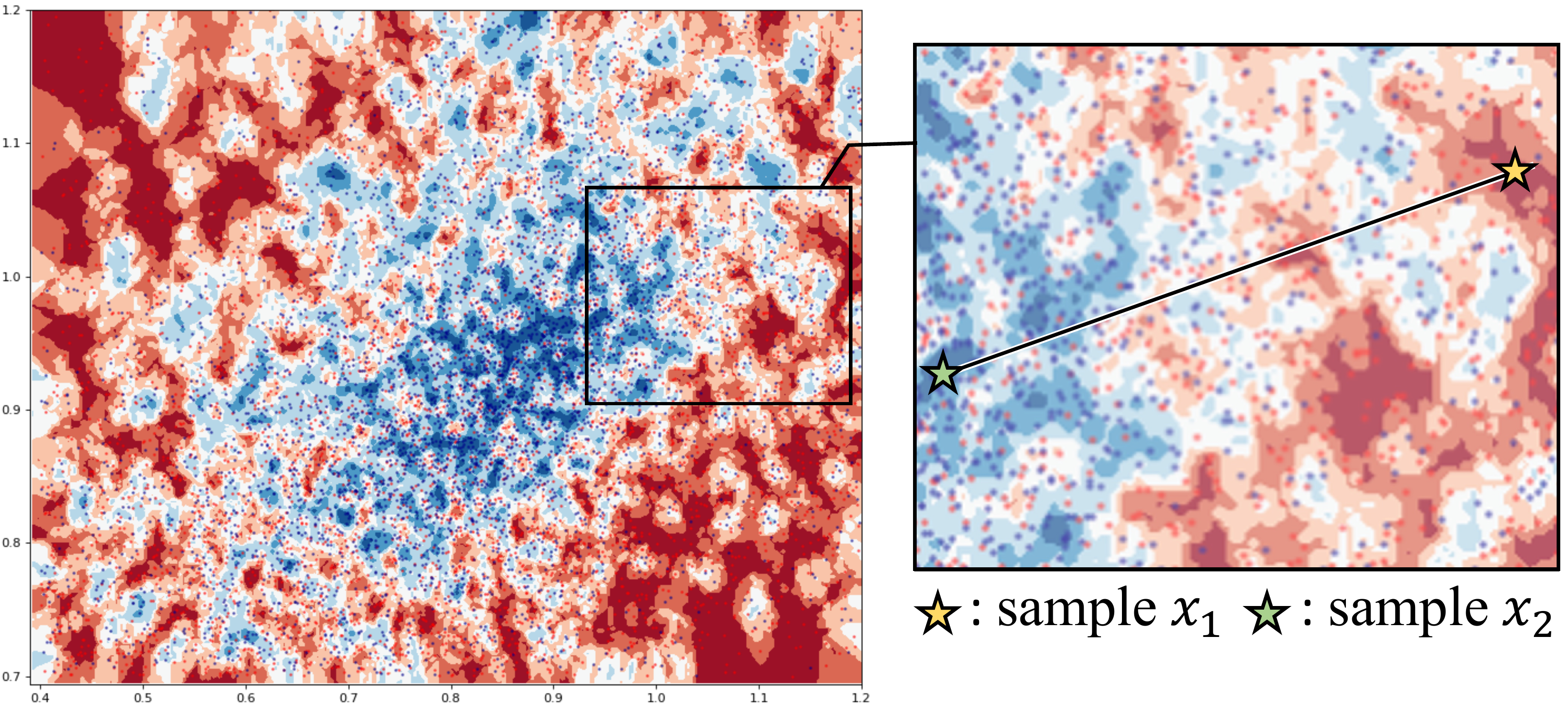}
    \vskip -1 em
    \caption{$k$NN ($k=8$) decision boundaries with 2 key features of a zoomed-in part of the Higgs dataset. Convex combinations by vanilla Mixup (points on the black line) of 2 samples $x_1$ and $x_2$ may conflict with irregular category boundaries.}\label{fig:pre-exp}
        \vskip -1.5 em
\end{figure}
As stated in \cite{ng2004feature}, less informative features make minor contributions on target prediction but still necessitate at least a linear increase in the requirement for training samples to learn how to ``ignore'' them. DNNs are rotationally invariant algorithms, which are data-inefficient with a worst-case sample complexity increasing at least linearly with the number of uninformative features \citep{grinsztajntree}.

Our idea is to incorporate an inductive bias into the self-attention mechanism, which selectively restricts the impacts of a feature to only those that are less informative, thereby reducing the overall impact of uninformative features on prediction outcomes. We propose a \textit{semi-permeable attention} module (SPA), as:
\begin{equation}\label{eq:sa}
    z^\prime = \text{softmax}(\frac{(zW_q) (zW_k)^T \oplus M}{\sqrt{d}}) (zW_v),
\end{equation}
where 
$z \in \mathbb{R}^{f \times d}$ is the input embeddings and $z^\prime$ the output embeddings,
$W_q, W_k$, $W_v \in \mathbb{R}^{d \times d}$ are all learnable matrices, and $\oplus$ is element-wise addition. $M\in \mathbb{R}^{f\times f}$ is an unoptimizable mask, where the element at the $i$-th row and $j$-th column is defined by:
\begin{equation}
M[i, j] =
\begin{cases}
-\infty & I(\mathbf{f}_i) > I(\mathbf{f}_j) \\
0 & I(\mathbf{f}_i) \leq I(\mathbf{f}_j)
\end{cases}
\end{equation}
The function $I(\cdot)$ represents a measure of feature importance, and we use the ``mutual information'' metric in this paper (see Appendix~\ref{sec:metric} for details). If a feature $\mathbf{f}_i$ is more informative compared to $\mathbf{f}_j$, $M[i, j]$ is set $-\infty$ (we use $-10^{5}$ in implementation) and thus the $(i,j)$ grid on the attention map is masked. It prevents the transfer of the embedding of the feature $\mathbf{f}_j$ to the $\mathbf{f}_i$'s embedding.

In this way, only more informative features are permitted to propagate information to the less informative ones, and the reverse is not allowed. By doing so, SPA still maintains interaction pathways between any two features while constraining the impacts of less informative ones. Intuitively, when training samples are insufficient, some feature interactions conducted by the model may be sub-optimal, as vanilla self-attention was proved data-inefficient~\cite{touvron2021training}. When using SPA, it can avoid the excessive impacts of a noisy feature on prediction outcomes in case some associated interaction pathways are ill-suited. 
Furthermore, the SPA inhibits certain feature transfer pathways, thereby obviating the need for the \model to learn partial rotational directions. The rotational properties of the \model lie between those of the rotational invariant counterpart with vanilla self-attention (e.g., FT-Transformer) and a fully rotational variant model (e.g., feedforward network conducting no feature interactions). Namely, our SPA partially disrupts DNNs' rotational invariance property.
In practice, SPA is extended to a multi-head self-attention version, with 32 heads by default.
\begin{figure}[t]
    \centering
    \includegraphics[width=0.4\textwidth]{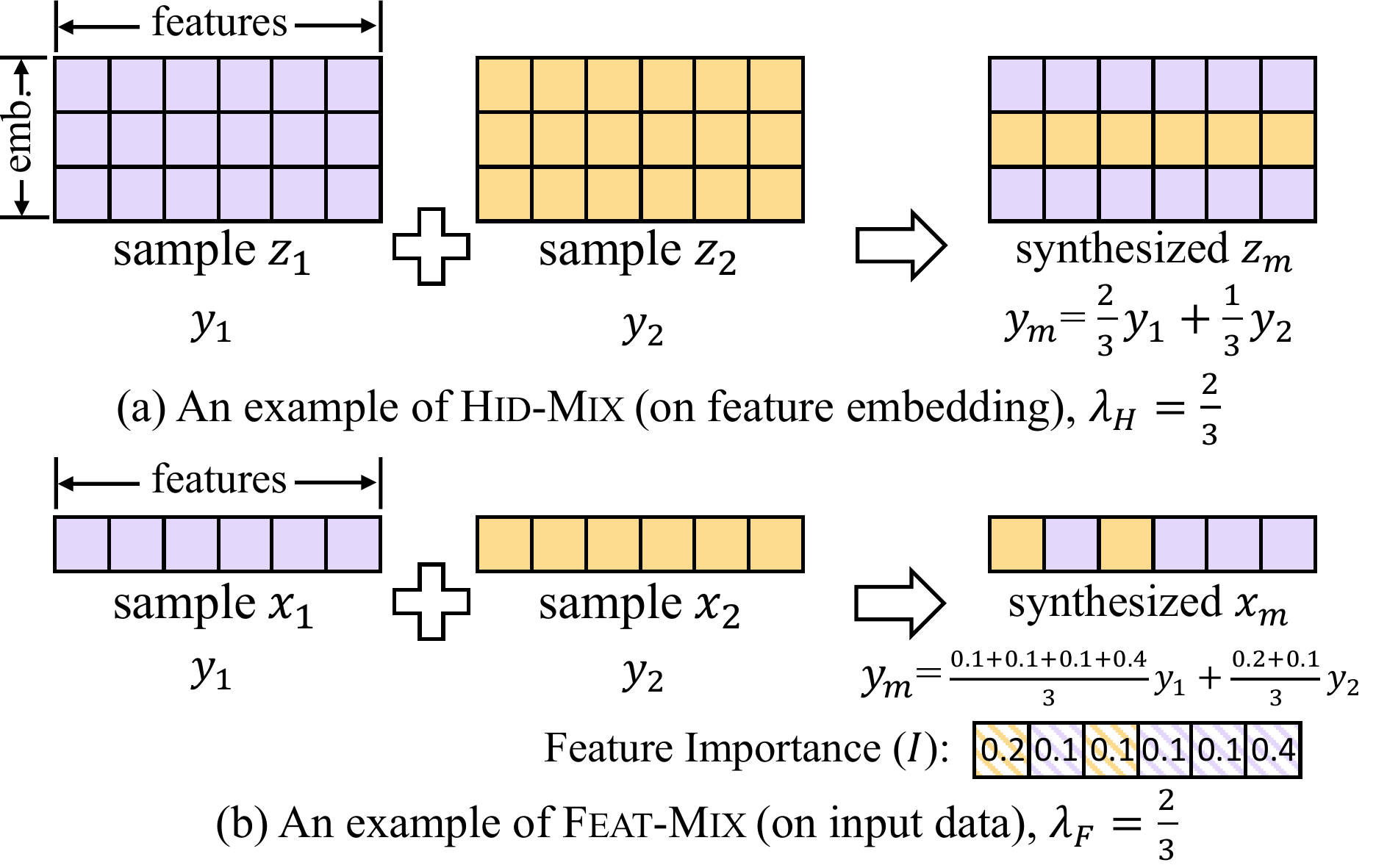}
    \vskip -1 em
    \caption{Examples for the \hidmix and \featmix, where ``emb.'' means ``embeding'' dimension.}\label{fig:mixup}
    \vskip -2 em
\end{figure}
\paragraph{\textbf{Interaction Attenuated Initialization.}} Similar to how the SPA disrupts rotational invariance by diminishing feature interactions,
we present a specialized initialization approach for SPA to ensure that \model starts as a largely non-rotationally invariant model. Notably, removing all self-attention operations from a Transformer model, features are processed individually, which makes the Transformer model nearly non-rotationally invariant (if we set aside the full connection layers that fuse features for target prediction). Concurrently, prior researches have evidenced the indispensable role of feature interactions (e.g., through self-attention) in Transformer-based models on tabular data~\cite{gorishniy2021revisiting,T2G}. By integrating these insights, our proposed \textit{interaction attenuated initialization} scheme initially dampens the impact of SPA during the early stages of training, allowing essential feature interactions progressively grow under the driving force of the data.


Our \textit{interaction attenuated initialization} scheme is built upon the commonly used He's initialization~\cite{he2015delving} or \textit{Xavier} initialization~\cite{glorot2010understanding}, by rescaling the variance of an initialized weight $w$ with $\gamma$ ($\gamma \rightarrow 0^+$) while keeping the expectation at 0:
\begin{equation}\label{eq:init}
    \text{Var}(w) = \gamma \text{Var}_{\text{prev}}(w),
\end{equation}
where $\text{Var}_{\text{prev}}(w)$ denotes the weight variance used in the He's initialization and \textit{Xavier} initialization. In this work, we set $\gamma=10^{-4}$. To reduce the impacts of SPA, 
we apply Eq.~(\ref{eq:init}) to all the parameters in the SPA module. Thus, \model works like a non-rotationally invariant model initially.

Actually, for a module with an additive identity shortcut like $y=\mathcal{F}(x) + x$, our initialization approach attenuates the sub-network $\mathcal{F}(x)$ and satisfies the property of \textit{dynamical isometry}~\cite{saxe2013exact} for better trainability. Some previous work~\cite{bachlechner2021rezero,touvron2021going} suggested to rescale the $\mathcal{F}(x)$ path as $y=\eta\mathcal{F}(x) + x$, where $\eta$ is a learnable scalar initialized as 0 or a learnable diagonal matrix whose elements are of very small values. Different from these methods, our attenuated initialization approach directly assigns minuscule values to the weights during initialization. Our approach is better suited for the flexible learning of whether each feature interaction pathway should be activated or not, thereby achieving sparse attention.

\vskip -0.8 em
\subsection{Solving (P3) with GLU layer}\label{sec:architecture}
The irregularities (Fig.~\ref{fig:pre-exp} shows an example) present in the tabular feature-target relationship make it particularly advantageous for decision trees utilizing multiple thresholds to split the feature space. 
On the contrary, existing Transformer employs a two-layer MLP as its feedforward network (FFN), which possesses a lesser degree of non-linear fitting capability. Therefore, we replace the vanilla FFN by a Gated Linear Unit (GLU) layer. Diverging from the standard GLU architecture, we employ the ``tanh'' activation in lieu of the ``sigmoid'' activation for better optimization properties~\cite{lecun2002efficient}, as:
\begin{equation}\label{eq:aium}
    z^\prime = \tanh{(\texttt{Linear}_1(z))} \odot \texttt{Linear}_2(z),
\end{equation}
where $\texttt{Linear}_1$ and $\texttt{Linear}_2$ are applied onto the embedding dimension $d$ of $z$,
$\odot$ denotes element-wise product. Please note that both the vanilla FFN and GLU employ two fully connection layers (FFN is defined by $z^\prime = \texttt{Linear}_1(\text{ReLU}(\texttt{Linear}_2(z)))$), resulting in similar computational costs. The SPA and GLU modules are alternately stacked to form the core structure of the \model model, as shown in Fig.~\ref{fig:framework}.

Existing tabular Transformers~\citep{T2G,gorishniy2021revisiting} use linear embedding layer to independently deal with each feature $\mathbf{f}_i \in \mathbb{R}$ into an embedding $\mathbf{z}_i \in \mathbb{R}^{d}$, by $z_i = \mathbf{f}_i W_{i,1} + b_{i,1}$. Here we also use GLU to replace it by $\mathbf{z}_i = \tanh{(\mathbf{f}_i W_{i,1} + b_{i,1})} \odot (\mathbf{f}_i W_{i,2} + b_{i,2})$, where $W_{i,1}, W_{i,2} \in \mathbb{R}^{1\times d}$ and $b_{i,1}, b_{i,2} \in \mathbb{R}^{d}$ are learnable parameters. Then, the initial feature embedding $z^{(0)}$ are obtained by stacking all $\mathbf{z}_i (i=1,2,\ldots,f)$, as $z^{(0)}=[\mathbf{z}_1, \mathbf{z}_2, \mathbf{z}_3, \ldots, \mathbf{z}_f]^T \in \mathbb{R}^{f\times d}$ like previous work.
\subsection{The rest part of \model}
\paragraph{\textbf{Feature Pre-processing.}} Feature values are pre-processed before feeding into \model. The numerical features are normalized by quantile transformation and the categorical features are converted into numerical ones using the CatBoost Encoder implemented with the \textit{Sklearn} Python package~\footnote{\url{https://contrib.scikit-learn.org/category_encoders/catboost.html}}. This step performs similar to previous works (e.g., FT-Transformer \cite{gorishniy2021revisiting}).

\paragraph{\textbf{Prediction Head.}}
The prediction head is directly applied to the output of the topmost transformer block, which contains two fully connection layers to separately compress the information along the token embeddings and fuse the information from features, by:
\begin{equation}
    p = \phi(\texttt{Linear}_d(\text{P-ReLU}(\texttt{Linear}_f(z^{(L)})))),
\end{equation}
where $z^{(L)}$ is the input, $W_f \in \mathbb{R}^{f\times C}$ and $b_f \in \mathbb{R}^C$. For multi-classification task, $C$ is the amount of target categories and $\phi$ indicates ``softmax''. For regression and binary classification tasks, then $C=1$ and $\phi$ is \textit{sigmoid}. The fully connection layer $\texttt{Linear}_f$ and $\texttt{Linear}_d$ are applied along and the feature dimension and the embedding dimension of $z^{(L)}$, respectively.
\section{Solving (P2) with Data Augmentation}\label{sec:mixupall}
A straightforward approach to tackle data insufficiency is to create training data. While Mixup~\cite{zhang2018mixup} regularizes DNNs to favor linear behaviors between samples and stands as one of the most effective data augmentation methods in computer vision, empirical evidence suggests that it does not perform optimally on tabular datasets (e.g., see Table~\ref{tab:dataaug}). 
This discrepancy may be due to the conflict between the model's linear behavior and the irregularity of target functions, as intuitively illustrated in Fig.~\ref{fig:pre-exp}.
To address this challenge, we introduce two Mixup variants, \hidmix and \featmix, which mitigate the conflicts in creating samples.

\textbf{\hidmix.} Our \hidmix is applied to the token-level embeddings after the input samples have been processed by the embedding layer, along with their corresponding labels. It randomly exchanges some embedding ``dimensions'' between two samples (please refer to Fig.~\ref{fig:mixup}(a)). Let $z_1^{(0)}, z_2^{(0)} \in \mathbb{R}^{f \times d}$ be the token-level embeddings of two randomly selected samples, with $y_1$ and $y_2$ denoting their respective labels. A new sample represented as a token-label pair $(z_{\text{m}}^{(0)}, y_{\text{m}})$ is synthesized by: 
\begin{equation}
\left\{
\begin{aligned}
    & z_{\text{m}}^{(0)}= S_H\odot z_1^{(0)} + (\mathbbm{1}_H - S_H) \odot z_2^{(0)}, \\
    & y_{\text{m}} = \lambda_H y_1 + (1-\lambda_H) y_2,
\end{aligned}
\right.
\end{equation}
where the matrix $S_H$ is of size $f \times d$ and is formed by stacking $f$ identical $d$-dimensional binary vectors denoted as $s_h$: $S_H = [s_h,s_h,\ldots, s_h]^T$. $s_h$ consists of $\lfloor \lambda_H \times d \rfloor$ randomly selected elements set to 1 and the rest elements set to 0. The scalar coefficient $\lambda_H$ for labels is sampled from the $\mathcal{B}eta(\alpha_H,\alpha_H)$ distribution, where $\alpha_H$ is a hyper-parameter. $\mathbbm{1}_H$ is an all-one matrix with dimensions $f \times d$. In practice, $\lambda_H$ is first sampled from given $\mathcal{B}eta(\alpha_H,\alpha_H)$ distribution. Subsequently, we randomly select $\lfloor \lambda_H \times d \rfloor$ elements to construct the vector $s_h$ and the matrix $S_H$.

Since the embedding ``dimensions'' from different samples may be randomly combined in training, \model is encouraged to independently and equally handle various embedding dimensions. Considering each embedding dimension as a distinct ``profile'' version of input data (as each embedding element is projected from a scalar feature value), \hidmix regularizes \model to behave like a bagging predictor~\cite{breiman1996bagging}. Therefore, \hidmix may also help mitigate the effects of data noise and perturbations, in addition to increasing the amount of training data.

\textbf{\featmix.} Our idea of \featmix is visualized as in Fig.~\ref{fig:mixup}. Unlike \hidmix that operates on the embedding dimension, our \featmix synthesizes new sample $(x_m, y_m)$ by swapping parts of features between two randomly selected samples $x_1,x_2 \in \mathbb{R}^f$, and blending their labels $y_1$ and $y_2$ guided by feature importance, by:
\begin{equation}
\left\{
\begin{aligned}
    & x_{\text{m}}= \mathbf{s}_F \odot x_1 + (\mathbbm{1}_F - \mathbf{s}_F) \odot x_2, \\
    & y_{\text{m}} = \Lambda y_1 + (1-\Lambda) y_2,
\end{aligned}
\right.
\end{equation}
where 
the vector $\mathbf{s}_F$ and 
the all-one vector $\mathbbm{1}_F$ are of size $f$, $\mathbf{s}_F$ contains $\lfloor \lambda_F \times f\rfloor$ randomly chosen elements set to 1 and the remaining elements set to 0. $\lambda_F \sim \mathcal{B}eta(\alpha_F,\alpha_F)$.
The coefficient value, $\Lambda$, is determined based on the contribution of $x_1$ and $x_2$, taking into account feature importance, by:
\begin{equation}
    \Lambda = \frac{\sum_{\mathbf{s}_F^{(i)}=1}I(\mathbf{f}_i)}{\sum^f_{i=1} I(\mathbf{f}_i)},
\end{equation}
where $\mathbf{s}_F^{(i)}$ represents the $i$-th element of $\mathbf{s}_F$, and $I(\cdot)$ returns the feature importance using mutual information. When disregarding feature importance, $\Lambda = \lambda_F$ (assuming $\lfloor \lambda_F \times f\rfloor = \lambda_F \times f$), making \featmix degenerate into a form similar to cutmix~\cite{yun2019cutmix}. However, due to the presence of uninformative features in tabular datasets, \featmix emerges as a more robust scheme.

As features from two distinct samples are randomly combined to create new samples, \featmix promotes a solution with fewer feature interaction. This aligns with the functionality similar to our Interaction Attenuated Initialization (see Sec.~\ref{sec:diam}). We argue that \featmix not only supplements the training dataset as a data augmentation method, but also encourages \model to predominantly exhibit like a non-rotationally invariant algorithm.
\section{Training and Loss Functions}
\model can handle both classification and regression tasks on tabular datasets in supervised learning. In training, our two proposed data augmentation schemes can be applied successively by $\hidmix(\text{Embedding Layer}(\featmix(x, y)))$ or used independently. But, our tests suggest that the effect of \model on a certain dataset could be better by using only \featmix or \hidmix. Thus, we use only one scheme in dealing with certain tabular datasets. The cross-entropy loss is used for classification tasks, and the mean square error loss is for regression tasks.
\begin{table*}[t]
\caption{Performance ranking ($\downarrow$) across 96 small tabular datasets containing fewer than 10k samples. Each model underwent 5 independent trials, with the model's average rank ($\pm$ std) reported. The best ranks are highlighted in \textbf{bold} while the runners-up are \underline{underlined}. Our \model consistently outperforms prior methods that undergo hyperparameter fine-tuning, regardless of whether \model uses fine-tuned or default hyperparameters. ``d'': using default hyper-parameters; ``t'': using tuned hyperparameters; ``No DA'': neither \featmix nor \hidmix is used.}\label{tab:small}
\vskip -0.9 em
\begin{tabular}{l|c|c|c|c|c}
\toprule
\multicolumn{1}{c}{\model setting:} & \multicolumn{1}{c}{No DA (t)} & \multicolumn{1}{c}{\featmix (d)} & \multicolumn{1}{c}{\hidmix (d)} & \multicolumn{1}{c}{Mix Tuned (t)} & \multicolumn{1}{c}{Fully Tuned (t)} \\
\midrule
XGboost (t)                        & \underline{4.20} $\pm$ 2.76  & \underline{4.21} $\pm$ 2.70 & \underline{4.29} $\pm$ 2.73 & \underline{4.34} $\pm$ 2.73 & \underline{4.28} $\pm$ 2.77 \\
Catboost (t)                    & 4.61 $\pm$ 2.73    & 4.57 $\pm$ 2.69 & 4.63 $\pm$ 2.68 & 4.66 $\pm$ 2.61 & 4.64 $\pm$ 2.68 \\ \midrule
FTT (t)                &  4.32 $\pm$ 2.36  & 4.35 $\pm$ 2.35 & 4.41 $\pm$ 2.25 & 4.44 $\pm$ 2.32 & 4.39 $\pm$ 2.37 \\
MLP (t)                        & 5.23 $\pm$ 2.31     & 5.27 $\pm$ 2.34 & 5.26 $\pm$ 2.32 & 5.30 $\pm$ 2.37 & 5.32 $\pm$ 2.33 \\
DCN v2 (t)                     & 6.01 $\pm$ 2.78     & 5.96 $\pm$ 2.75 & 5.99 $\pm$ 2.27 & 6.03 $\pm$ 2.74 & 6.02 $\pm$ 2.73 \\
AutoInt (t)                &  5.70 $\pm$ 2.61     & 5.78 $\pm$ 2.51 & 5.77 $\pm$ 2.56 & 5.88 $\pm$ 2.53 & 5.80 $\pm$ 2.55 \\
SAINT (t)               &  5.48 $\pm$ 2.59      & 5.48 $\pm$ 2.55 & 5.56 $\pm$ 2.56 & 5.61 $\pm$ 2.55 & 5.56 $\pm$ 2.58 \\ 
TransTab (d)                &  6.78 $\pm$ 2.52    & 6.80 $\pm$ 2.59 & 6.82 $\pm$ 2.57 & 6.86 $\pm$ 2.59 & 6.87 $\pm$ 2.55 \\
XTab (d)                    &  8.56 $\pm$ 2.20     & 8.68 $\pm$ 2.19 & 8.67 $\pm$ 2.19 & 8.67 $\pm$ 2.19 & 8.71 $\pm$ 2.14 \\
\midrule
\model (ours)                    & \textbf{4.11} $\pm$ 2.68   & \textbf{3.91} $\pm$ 2.60 & \textbf{3.62} $\pm$ 2.59 & \textbf{3.20} $\pm$ 2.10 & \textbf{3.41} $\pm$ 2.12 \\
\bottomrule
\end{tabular}
\end{table*}
\section{Experiments}\label{sec-Exp}
\subsection{Experimental Setups}\label{sec:expsetup}
\textbf{Implementation Details.} We configure the number of SPA and GRU modules as $L=3$, set the feature embedding size to $d=256$, and apply a dropout rate of 0.3 to the attention map. AdamW optimizer~\citep{loshchilov2018decoupled} is used with default settings. The learning rate is set to $10^{-4}$ without weight decay, and $\alpha_H$ and $\alpha_F$ for $\mathcal{B}eta$ distributions are both set to 0.5. These settings are the default hyperparameters for our \model.
In the hyperparameter fine-tuning process, we utilized the Optuna library~\citep{akiba2019optuna} for all approaches. Consistent with~\citep{gorishniy2021revisiting}, we randomly select 80\% of the data as training samples and the remaining 20\% as test samples. During training, we reserve 20\% training samples for validation. To fine-tune our \model, we designate two tuning configurations: ``Mix Tuned'' and ``Fully Tuned''. ``Mix Tuned'' refers to only fine-tune hyperparameters of data augmentation (for \featmix and \hidmix), while ``Fully Tuned'' optimizes all hyperparameters, including those related to data augmentation and model architecture. A comprehensive description of all settings can be found in Appendix~\ref{app:tune}. We applied early stopping with a patience of 32 for \model.

\textbf{Datasets.} A total of 96 small datasets sourced from the Taptap dataset benchmark\footnote{\url{https://huggingface.co/datasets/ztphs980/taptap_datasets}} were utilized. The criterion for classifying datasets as small is based on having a sample size of less than 10,000.
Besides, 21 larger public tabular datasets, ranging in scale from over 10,000 to 581,835 samples were also used. The detailed dataset descriptions are provided in Appendix~\ref{app:datasets}.

\textbf{Compared Models.} We compare our new \model with two prominent GBDT approaches XGboost~\citep{chen2016xgboost} and Catboost~\citep{prokhorenkova2018catboost} and several representative DNNs: FT-Transformer (FTT)~\citep{gorishniy2021revisiting}, SAINT~\citep{somepalli2021saint}, Multilayer Perceptron (MLP), DCN v2~\citep{wang2021dcn}, AutoInt~\citep{song2019autoint}, and TapPFN \citep{tabpfn}. We also include two pre-trained DNNs: TransTab~\citep{wang2022transtab} and XTab~\citep{zhu2023xtab} for reference. The implementations of XGboost and Catboost mainly follow~\citep{gorishniy2021revisiting}. Since we aim to extensively tune XGboost and Catboost for their best performances, we increase the maximum number of estimators/ iterations (i.e., the number of decision trees) from 2000 to 4096 and the number of tuning iterations from 100 to 500, which give a more stringent setting and better performances. The settings for XGboost and Catboost are given in Appendix~\ref{app:tune}. We use the default hyperparameters of pretrained models, TransTab and XTab, and fine-tune them on each dataset. They are not hyperparameter tuned, since their hyperparameter tuning spaces are not given. For large-scale datasets, FT-Transformer, SAINT, and TapFPN were fine-tuned based on the hyperparameters outlined in their respective papers. The architectures and hyperparameter tuning settings of the remaining DNNs follows the paper~\citep{gorishniy2021revisiting}. On small datasets, we tuned 50 iterations for each datasets.

\textbf{Evaluation metrics.} We use the area under the ROC Curve (AUC) and accuracy (ACC) for binary classification tasks and multi-class classification tasks. In regression tasks, we employ the negative root mean square error (nRMSE), where the negative sign is introduced to RMSE, aligning its direction with AUC and ACC, such that higher values across all these metrics indicate superior performance. Due to the high diversity among tabular datasets, performance ranks are used as a comprehensive metric, and the detailed results are given in Appendix~\ref{app:performances}. 
\begin{table}[t]
\caption{Performance evaluation across 21 larger-scale datasets, each containing more than 10,000 samples, is conducted. Average ranks with standard deviations are reported based on the results of 5 runs with different random seeds. \textsc{Excel} defines \model. The best and second best performances are \textbf{bold} and \underline{underlined}.}\label{tab:large}
\vskip -0.9 em
\begin{tabular}{c|l|c}
\hline
Setting  & \multicolumn{1}{c|}{Model} & \multicolumn{1}{c}{Rank (mean $\pm$ std)} \\ \hline
\multirow{3}{*}{default} & XGboost             & 8.52 $\pm$ 1.86                          \\
\multirow{3}{*}{hyperparameters}                         & Catboost             & 7.52 $\pm$ 2.44                      \\
                         & FTT         & 6.71 $\pm$ 1.74                           \\
                         & Excel w/ \featmix   & \underline{6.62} $\pm$ 2.44                      \\
                         & Excel w/ \hidmix    & \textbf{4.76} $\pm$ 1.95                    \\ \hline
\multirow{3}{*}{hyperparameter}   & XGboost        & 4.29 $\pm$ 2.59                     \\
                         & Catboost         & 6.24 $\pm$ 2.39                     \\
\multirow{2}{*}{fine-tuned}       & FT-T   & 5.19 $\pm$ 2.60                  \\
                         & Excel (Mix Tuned)                 & \underline{2.38} $\pm$ 1.53                 \\
                         & Excel (Fully Tuned)               & \textbf{2.05} $\pm$ 1.40                 \\ \hline
\end{tabular}
\vskip -1.5 em
\end{table}
\begin{table*}[t]
\caption{Performance evaluation within several dataset subgroups. Performance rank ($\downarrow$) within the datasets are reported. 
The best scores are in \textbf{bold} and the runners-up are \underline{underlined}. ``(d)'': default hyperparameters; ``(t)'': finely tuned hyperparameters. 
}\label{tab:multihead}
\vskip -0.7 em
\setlength{\tabcolsep}{7pt}
\resizebox{\textwidth}{!}{
\begin{tabular}{l|ccccccccccc}
\toprule
\multicolumn{1}{c}{Model}      & \model & FTT (t) & XGb (t) & Cat (t) & MLP (t)  & DCNv2 (t) & AutoInt (t) & SAINT (t) & TransTab (d) & XTab (d) & TabPFN (t) \\ \hline
\rowcolor{camel} Characteristics: Task Type     & \multicolumn{11}{c}{Classification}                                                                                                                      \\
\rowcolor{lightgray} Proportion                     & \multicolumn{11}{c}{51\%}                                                                                                                                \\
Setting: \hidmix (d)     & \textbf{3.88}   & 4.88    & 5.97       & 5.77   & 6.61 & 6.38  & 6.63   & 6.07 & 6.31       & 9.50   & \underline{4.01}                       \\
Setting: Mix Tuned       & \textbf{3.78}   & 4.91   & 5.95   & 5.79  & 6.60 & 6.39 & 6.71 & 6.10 & 6.37    & 9.46    & \underline{3.95}                       \\
\rowcolor{camel} Characteristics: Task Type     & \multicolumn{11}{c}{Regression}     \\
\rowcolor{lightgray} Proportion                     & \multicolumn{11}{c}{49\%}                                                                                                                                \\
Setting: \hidmix (d)     & \underline{3.81}   & 4.45    & \textbf{3.43}    & 4.26   & 4.64 &  6.26   & 5.53  & 5.64  & 8.21 & 8.79     & /                        \\
Setting: Mix Tuned       & \textbf{3.17}   & 4.49   & \underline{3.53}        & 4.28  & 4.74 &  6.32   & 5.68   & 5.72 & 8.23     & 8.83                 & /                        \\ \hline
\rowcolor{camel}Characteristics: \#. Sample & \multicolumn{11}{c}{$\geq$ 500}                                                                                                                          \\
\rowcolor{lightgray} Proportion                     & \multicolumn{11}{c}{43\%}                                                                                                                                \\
Setting: \hidmix (d)     & \textbf{3.85}   & 4.50    & \underline{4.38}        & 5.17         & 5.57 & 5.91   & 5.59    & 5.24  & 6.44        & 8.34                        & /                        \\
Setting: Mix Tuned             & \textbf{3.52}   & 4.50    & \underline{4.39}        & 5.15         & 5.60 & 5.99   & 5.71    & 5.33  & 6.48        & 8.34                        & /                        \\
\rowcolor{camel} Characteristics: \#. Sample & \multicolumn{11}{c}{< 500}                 \\
\rowcolor{lightgray} Proportion                     & \multicolumn{11}{c}{57\%}      \\
Setting: \hidmix (d)     & \textbf{3.45}   & 4.34    & \underline{4.22}        & 4.23         & 5.02 & 6.05   & 5.90    & 5.79  & 7.10        & 8.92                        & /                        \\
Setting: Mix Tuned             & \textbf{3.18}   & 4.38    & \underline{4.28}        & 4.29         & 5.05 & 6.05   & 5.97    & 5.79  & 7.13        & 8.88                        & /                        \\ \hline
\rowcolor{camel} Characteristics: \#. Feature   & \multicolumn{11}{c}{\#. Feature < 8}          \\
\rowcolor{lightgray} Proportion                     & \multicolumn{11}{c}{32\%}              \\
Setting: \hidmix (d)     & \textbf{3.45}   & \underline{3.84}    & 3.98        & 5.08         & 4.23 & 6.32   & 6.16    & 5.32  & 7.52        & 9.10                        & /                        \\
Setting: Mix Tuned             & \textbf{3.27}  & \underline{3.84}    & 4.03        & 5.06         & 4.34 & 6.26   & 6.21    & 5.35  & 7.50        & 9.13                        & /                        \\
\rowcolor{camel} Characteristics: \#. Feature   & \multicolumn{11}{c}{8 $\leq$ \#. Feature < 16 }          \\
\rowcolor{lightgray} Proportion                     & \multicolumn{11}{c}{38\%}         \\
Setting: \hidmix (d)     & \textbf{3.76}   & \underline{4.26}    & 4.44        & 4.61         & 6.31 & 5.75   & 5.39    & 5.69  & 6.61        & 8.17                        & /                        \\
Setting: Mix Tuned             & \textbf{3.17}   & \underline{4.33}    & 4.49        & 4.74         & 6.33 & 5.81   & 5.58    & 5.78  & 6.64        & 8.14                        & /                        \\
\rowcolor{camel} Characteristics: \#. Feature   & \multicolumn{11}{c}{\#. Feature $\geq$ 16 }                      \\
\rowcolor{lightgray} Proportion                     & \multicolumn{11}{c}{30\%}                             \\
Setting: \hidmix (d)     & \textbf{3.62}   & 5.19    & 4.41        & \underline{4.17}         & 5.05 & 5.93   & 5.81    & 5.64  & 6.33        & 8.84                        & /                        \\
Setting: Mix Tuned             & \textbf{3.17}   & 5.22    & 4.48        & \underline{4.14}        & 5.05 & 6.05   & 5.90    & 5.69  & 6.45        & 8.84                        & /                        \\ \hline
\end{tabular}}
\vskip -1.0 em
\end{table*}
\subsection{Model Performances and Discussions}\label{sec:resultanddiscussion}
\paragraph{\textbf{Performances on Small Datasets.}} DNNs are typically data-inefficient, thus we initially investigate whether the proposed \model can effectively perform on small datasets. See Table~\ref{tab:small}, our \model consistently outperforms other models that undergo \textbf{dataset-adaptive hyperparameter tuning}, regardless of whether the hyperparameters of the \model are tuned or not, which underscores the superiority of our proposed \model. 
We observe that \model with \hidmix slightly outperforms that with \featmix; and if we tune hyperparameters of \model, its performance achieves further improvement. Notably, hyperparameter fine-tuning reduces the standard deviations of performance ranks, indicating that applying hyperparameter fine-tuning onto \model can yield more consistently superior results. Interestingly, while fine-tuning all the hyperparameters (``Fully Tuned'') should result in better performance ideally, it shows that, under the same fine-tuning iterations, ``Mix Tuned'' configuration performs better. This might be attributed to the higher efficiency of finely tuning data augmentation setting. To assess the effectiveness of our \model's architecture, we conducted experiments by excluding all data augmentation (\featmix and \hidmix) and compare it with existing models. The results show that even without the use of \featmix and \hidmix, \model still outperforms previous approaches, underscoring the superiority of our architecture.

\paragraph{\textbf{Performances on Larger Datasets.}} We further conduct a comparison between our model and three previous state-of-the-art models: XGboost, Catboost, and FTT. We excluded other models from the comparison due to their relatively inferior performances and the significant computational load when dealing with large datasets. Each model undergoes evaluation with two settings: using default hyperparameters and dataset-adaptive fine-tuning hyperparameters. As depicted in Table~\ref{tab:large}, our model also outperforms the previous models under both settings. Additionally, it is worth noting that our \model with \hidmix still achieves comparable performance to prior models that undergo hyperparameter tuning, consistent with the findings on small datasets. Different from the conclusion on small datasets, the Fully Tuned \model outperforms the Mix Tuned version on large datasets.

\textbf{Takeaway.} We discovered that \textbf{(1)} our model performs well on GBDT-favored smaller datasets and DNN-favored larger ones. This suggests that our design addresses the existing drawbacks of DNNs in tabular prediction. \textbf{(2)} Even with default hyperparameters, \model consistently outperforms hyperparameter-tuned competitors on small datasets and performs competitive with them on larger ones. This implies that for users who are not experts in hyperparameter tuning, using our model can still obtain a strong solution. Moreover, even for professional users, our model also stands out as a top choice since the hyperparameter-tuned \model performs excellent on various tabular prediction tasks.

\subsection{Can \model be a \textit{sure bet} solution across various types of datasets?}\label{3232}
\textit{We still have to further rigorously examine whether our model performs poorly on certain tabular dataset types to ensure that we have achieved our goal of building a \textit{sure bet} solution.}
We divide datasets into various subgroups according to the task type, dataset size, and the number of features, and examine \model performance within each subgroup. We adopt two configurations, \hidmix (default) and Mix Tuned, for \model, while \textbf{all of the existing models undergo hyperparameter fine-tuning.} As shown in Table~\ref{tab:multihead}, \model with \hidmix (default) exhibits the best performance in all subgroups except for regression tasks, where it slightly lags behind hyperparameter-tuned XGboost. The Mix Tuned \model outperforms other models in all subgroups, indicating that \model does not exhibit overt dataset type preferences. 

\textbf{Takeaway.} What changes does \model bring to the field of tabular data prediction? 
Refer to Table~\ref{tab:multihead}, besides our model, runner-up positions are held by TapFPN, FTT, XGBoost, and CatBoost in different subgroups. Notably, TapFPN is solely applicable to classification tasks, CatBoost performs well on datasets with numerous features, and FTT excels on datasets with fewer than 16 features. However, our proposed model demonstrates strong performance across all dataset types, which further proves its status as a \textit{sure be} solution for tabular datasets.

\subsection{Ablation Analysis}\label{3443}
\begin{table}[ht]
\caption{Additive study for the \textit{semi-permeable attention} (SPA) and \textit{interaction-attenuated initialization} (IAI), and the GLU based attentive module. No data augmentation.}\label{tab:additivestudy}
\vskip -1 em
\begin{tabular}{cccc|c}
\toprule
baseline & SPA & IAI & GLU & rank ($\pm$ std) \\ \midrule
\checkmark        &     &     &     & 4.31 $\pm$ 0.94 \\
\checkmark        &     & \checkmark   &     & 3.87 $\pm$ 1.58 \\
\checkmark        & \checkmark   &     &     & 3.73 $\pm$ 2.04 \\
\checkmark        & \checkmark   & \checkmark   &     & 2.45 $\pm$ 1.60 \\
\checkmark        &     &     & \checkmark   & 3.71 $\pm$ 1.52\\
\checkmark        & \checkmark   & \checkmark   & \checkmark   & \textbf{2.31} $\pm$ 1.46 \\ \bottomrule
\end{tabular}
\vskip -1 em
\end{table}
\begin{table}[htpb]
\caption{Comparison among several data augmentation approaches on FT-Transformer and \model. Ranks are computed separately for different backbones.}\label{tab:dataaug}
\vskip -1 em
\centering
\begin{tabular}{c|c|c}
\toprule
Backbone             & Data Augmentation & rank ($\pm$ std)\\ \midrule
\multirow{5}{*}{FT-Transformer} & N/A               & 3.28 $\pm$ 1.66 \\
                     & Mixup             & 3.80 $\pm$ 1.39\\
                     & CutMix            & 2.91 $\pm$ 1.37\\
                     & \featmix          & \underline{2.50} $\pm$ 1.03\\
                     & \hidmix           & \textbf{2.24} $\pm$ 1.00\\ \midrule
\multirow{5}{*}{\model} & N/A               & 3.68 $\pm$ 1.43\\
                     & Mixup             & 3.46 $\pm$ 1.63\\
                     & CutMix            & 2.88 $\pm$ 1.21\\
                     & \featmix          & \underline{2.38} $\pm$ 1.25\\
                     & \hidmix           & \textbf{2.36} $\pm$ 1.03\\ \bottomrule
\end{tabular}
\vskip -1 em
\end{table}
Here we investigate the effects of architectural design (SPA, GLU) and data augmentation approaches (\hidmix and \featmix), with the results presented in Table~\ref{tab:additivestudy} and Table~\ref{tab:dataaug}, respectively.

In Table~\ref{tab:additivestudy}, the baseline model employs an \model with the vanilla self-attention module initialized using typical Kaiming initialization~\cite{he2015delving}, along with a vanilla MLP-based FFN. Subsequently, we evaluate how the designed approaches enhance this baseline. It is observed that SPA and IAI individually improve baseline performances, and their joint usage achieves even better results. Additionally, GLU can also significantly enhances the baseline. These findings suggest that our architectural designs, SPA with IAI and GLU, are all well-suited for tabular data predictions. In the last row, where all these components are utilized (\model), we demonstrate that their combined utilization leads to the best results. The rotational invariance property brought by SPA and IAI are carefully demonstrated in Appendix~\ref{sec:rota}.

See Table~\ref{tab:dataaug}, where we report the comparison results among various data augmentation techniques on both the FTT backbone~\cite{rubachev2022revisiting} and our \model. It is crucial to recognize that the performance rankings are computed independently for different backbones, making direct comparisons of ranks unfeasible. It is evident that Mixup demonstrates minimal to no effect and sometimes even exhibits a detrimental impact. This could be attributed to Mixup's interpolation potentially introducing ``error'' cases or steering the model towards an overly smooth solution. In contrast, CutMix consistently outperforms Mixup, approaching the performance level of \featmix, albeit with slight inferiority. As discussed in Sec.~\ref{sec:mixupall}, without considering feature importance, \featmix may regress to CutMix; however, feature importance computation is crucial to mitigate the impacts of uninformative features, a common occurrence in tabular datasets. Further experiments are detailed in Appendix~\ref{sec:cutfeat}. It is evident that our proposed \featmix and \hidmix consistently enhance DNN model performance and prove more effective compared to other Mixup variants.
\vskip -1 em
\section{Conclusions}
This paper introduces a novel approach aimed at addressing three key limitations of DNNs when applied to tabular data prediction. We present a novel \textit{semi-permeable attention} module incorporated with an \textit{interaction attenuated} initialization approach, a GLU based FFN, as well as two data augmentation approaches: \hidmix and \featmix. Through the integration of these designs, we present \model, a model that maintains the same size as previous tabular Transformers but significantly outperforms existing GBDTs and DNNs in terms of performance, without hyperparameter tuning. Extensive and stratified experiments demonstrate that the \model stands out as a \textbf{\textit{sure bet}} solution for tabular prediction. We believe the proposed framework is highly accessible and user-friendly for even novices working with tabular data.

\noindent\textbf{Acknowledgements.} This work was mainly supported by NSF award SCH-2205289, SCH-2014438, and IIS-2034479.


\bibliographystyle{ACM-Reference-Format}
\bibliography{sample-base}

     
\appendix
\newpage
\section{Why is \featmix superior to CutMix?}\label{sec:cutfeat}
The primary distinction between the \featmix and CutMix approaches~\cite{yun2019cutmix} lies in whether the feature importance is considered when synthesizing new samples. To explore this difference, we conducted experiments on several datasets using the architecture of the \model as backbone. Our observations were made on both the original tables and the tables augmented with additional columns containing Gaussian noise. See Table~\ref{tab:noisy}, generally, \featmix outperforms CutMix or performs on par with CutMix on these datasets. However, in tables with noisy columns, we only observed a slight decline in the effectiveness of \featmix (even with an improvement on the cpu dataset), while CutMix exhibited a more significant performance drop under the influence of noisy columns. Given the prevalence of uninformative features in tabular data~\citep{ng2004feature,grinsztajntree}, the comparison of their performance and performance drops with noisy data emphasizes the importance of considering feature importance during interpolation. We find that \featmix stands out as a more resilient choice for tabular datasets. By leveraging \featmix instead of CutMix, \model can be deemed a more dependable approach for casual users, aligning with our initial intent in this research.
\begin{table*}[tbp]
\caption{Performance comparison between CutMix and \featmix. The first three datasets are for binary classification, with performance evaluated using the AUC ($\uparrow$). The rest datasets are for regression, assessed through nRMSE ($\uparrow$).``with noise'' indicates we add some noisy columns to the table. Breast: Breast Cancer Coimbra; Diabetes: Pima-Indians-Diabetes; Campus:Campus Recruitment; yacht: yacht\_hydrodynamics.}\label{tab:noisy}
\centering
\begin{tabular}{l|cccccc}
\toprule
                    & Breast & Diabetes & Campus & cpu    & fruitfly & yacht \\ \midrule
CutMix    & 0.702                 & 0.822                 & 0.972               & -102.06 & -16.19   & -3.59                \\
\featmix           & \textbf{0.713}             & \textbf{0.837}                 & \textbf{0.980}               & \textbf{-79.10}  & \textbf{-15.86}   & \textbf{-0.83}                \\ \midrule
CutMix (with noise)    & 0.688                  & 0.809                 & 0.938               & -115.10 & -17.09   & -4.40                \\
\featmix (with noise) & \textbf{0.700}                  & \textbf{0.834}                 & \textbf{0.969}               & \textbf{-74.56}  & \textbf{-16.60}   & \textbf{-0.89}               \\ \midrule
$\Delta$ CutMix ($\downarrow$)             & 0.014                  & 0.013                  & 0.034               & 13.04   & 0.90     & 0.81                 \\
$\Delta$ \featmix ($\downarrow$)             & 0.013                  & 0.003                  & 0.011               & -4.54   & 0.74     & 0.06        \\
\bottomrule
\end{tabular}
\end{table*}

\section{Rotational Variance Property Evaluation}\label{sec:rota}
\begin{figure*}
    \centering
    \includegraphics[width=\textwidth]{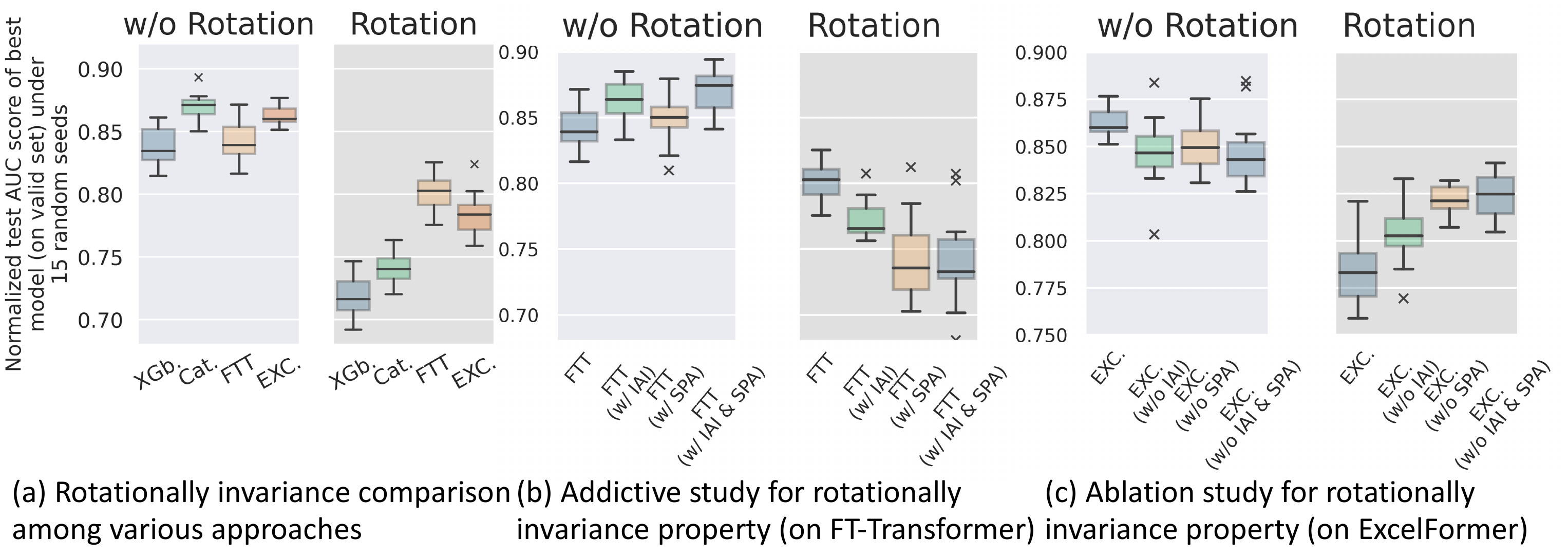}
    \vskip -1 em
    \caption{Model performances under random dataset rotations. Test accuracy scores have been normalized across datasets, and the boxes represent the distribution of scores across 20 random seeds. XGb.: XGboost, Cat.: Catboost, \textsc{EXC.}: \model without data augmentation.}\label{fig:rotation_boxplot}
\end{figure*}

In (P1) of main paper, we highlight the DNNs' lack of rotational variance, which makes them less efficient for tabular prediction tasks. This limitation serves as the motivation behind our proposal of the \textit{semi-permeable attention} (SPA) and the interaction attenuated initialization (IAI) approach.
Here we would like to inspect if the \model is more non-rotationally invariant with the proposed SPA and IAI.
We assess the test performance of \model (without using data augmentation) when randomly rotating the datasets. We utilize all binary classification datasets consisting of numerical features and containing fewer than 300 data samples. Additionally, we introduce $f$ uninformative features into each dataset (assuming that the original table comprises $f$ features), which are generated using Gaussian noises. As depicted in Fig.~\ref{fig:rotation_boxplot} (a), it is evident that after randomly rotating the datasets, XGBoost and CatBoost exhibit the most significant decline in performances. This observation suggests that they are algorithms with a higher degree of non-rotational invariance, aligning with the findings of \citep{ng2004feature}. While the decline in performance of \model and FTT are not as substantial as those of decision tree-based models, it is still noticeable that \model's performance decreases by a larger extent after random rotations, compared to FTT. This observation indicates that our \model exhibits a higher degree of non-rotational variance compared to counterpart FTT.

Inversely, we further conducted additive studies, utilizing FTT as the backbone and incorporating our proposed SPA and IAI on FTT. See Fig.~\ref{fig:rotation_boxplot}(b), we find that: (i) both SPA and IAI contribute positively to the performances of FTT. (ii) In the presence of random dataset rotations, FTT with IAI and SPA demonstrated a more pronounced performance drop, thereby showcasing the efficacy of SPA and IAI in enhancing the non-rotational invariance property of FTT. Additionally, see Fig.~\ref{fig:rotation_boxplot}(c), ablation studies on the \model backbone (where neither \featmix nor \hidmix was applied) also highlighted the value of SPA and IAI in mitigating the rotational invariance property of DNN models.
\begin{table*}[t]
\centering
\caption{Development time comparison between hyperparameter-tuned GBDTs and \model with default parameters.}\label{tab:time}
\begin{tabular}{l|ccccc|c}
\toprule
Model                 & eye     & california & house   & jannis  & higgs-small & average time (s)  \\ \midrule
XGboost (t)              & 699.92  & 325.48     & 238.95  & 2260.57 & 425.67      & 790.12 \\
Catboost (t)              & 1241.43 & 589.10     & 387.77   & 4100.39 & 714.53      & 1406.64 \\
\model (d)              & 39.29   & 51.80      & 59.41   & 166.57  & 160.48      & 95.51 \\ \bottomrule
\end{tabular}
\end{table*}
\begin{table*}[ht]
\centering
\caption{The hyper-parameter optimization space for \model. The items marked with ``*'' are used to obtain a ``Mix Tuned'' \model, while all the items are used to obtain a ``Fully Tuned'' version.}\label{tab:excel-tune}
\begin{tabular}{ll}
\toprule
Hyper-parameter & Distribution\\
\midrule
\#. Layers $L$ & UniformInt[2, 5]\\
Representation size $d$ & \{64, 128, 256\}\\
\#. Heads & \{4, 8, 16, 32\}\\
Residual dropout rate & Uniform[0, 0.5]\\
Learning rate & LogUniform[$3\times 10^{-5}$, $10^{-3}$]\\
Weight decay & \{0.0, LogUniform[$10^{-6}$, $10^{-3}$]\}\\ \midrule
(*) Mixup type & \{\featmix, \hidmix, neither\}\\
(*) $\alpha$ of $\mathcal{B}eta$ distribution & Uniform[0.1, 3.0]\\
\bottomrule
\end{tabular}
\end{table*}
\begin{table*}[ht]
\caption{The hyper-parameter tuning space for XGboost.}\label{tab:tune-xgboost}
\vskip -0.8 em
\begin{tabular}{ll}
\toprule
Hyper-parameter & Distribution\\ \midrule
Booster & ``gbtree'' \\
N-estimators & Const(4096) \\
Early-stopping-rounds & Const(50) \\
Max depth & UniformInt[3, 10] \\
Min child weight & LogUniform[$10^{-8}$, $10^{5}$] \\
Subsample & Uniform[0.5, 1.0] \\
Learning rate & LogUniform[$10^{-5}, 1$] \\
Col sample by level & Uniform[0.5, 1] \\
Col sample by tree & Uniform[0.5, 1] \\
Gamma & \{0, LogUniform[$10^{-8}, 10^2$]\} \\
Lambda & \{0, LogUniform[$10^{-8}, 10^2$]\}\\
Alpha & \{0, LogUniform[$10^{-8}, 10^2$]\}\\ \midrule
\#. Tuning iterations & 500\\
\bottomrule
\end{tabular}
\end{table*}
\begin{table*}[ht]
\caption{The hyper-parameter tuning space for Catboost.}\label{tab:tune-catboost}
\vskip -0.8 em
\begin{tabular}{ll}
\toprule
Hyper-parameter & Distribution\\ \midrule
Iterations (number of trees) & Const(4096) \\
Od pval & Const(0.001) \\
Early-stopping-rounds & Const(50) \\
Max depth & UniformInt[3, 10] \\
Learning rate & LogUniform[$10^{-5}, 1$] \\
Bagging temperature & Uniform[0, 1] \\
L2 leaf reg & LogUniform[1, 10] \\
Leaf estimation iterations & UniformInt[1, 10] \\\midrule
\#. Tuning iterations & 500\\
\bottomrule
\end{tabular}
\end{table*}
\section{Details of Hyper-Parameter Fine-Tuning Settings}\label{app:tune}
For XGboost and Catboost, we follow the implementations and settings in~\citep{gorishniy2021revisiting}, while increasing the number of estimators/iterations 
(i.e., decision trees) and the number of tuning iterations, so as to attain better performance. For our \model, we apply the Optuna based tuning~\citep{akiba2019optuna}. The hyper-parameter search spaces of \model, XGboost, and Catboost are reported in Table~\ref{tab:excel-tune}, Table~\ref{tab:tune-xgboost}, and Table~\ref{tab:tune-catboost}, respectively. For \model, we tune just 50 iterations on the configurations with regard to the data augmentation (it is marked as ``Mix Tuned''). For ``Fully Tuned'' version, we finely tune 50 interations on all the hyper-parameters.
\section{Development Time Cost}\label{app:time}
In the main paper, we have demonstrated on a large array of datasets that, with default hyperparameters, our proposed model outperforms existing models that require time-consuming heavy hyperparameter tuning (typically requiring 50-100 iterations). Among these, XGBoost and CatBoost stand out as two of the most efficient approaches. It is apparent in comparison to FTT that our model demands only approximately 1/100 to 1/50 of the development time. This is due to its close architecture of \model to FTT while being competitive or even superior without the need for hyperparameter tuning.

We conducted an analysis of the total time invested in model development for XGBoost, CatBoost, and our proposed \model. As shown in Table~\ref{tab:time}, we observed that our model can achieve significantly higher efficiency than both, even with just 50 iterations of hyperparameter tuning applied to XGBoost and CatBoost. Developing an XGBoost or CatBoost model requires 8-15 times more development time than our model, indicating that our approach is user-friendly and environmentally friendly.

\section{Implementation Details of Metrics Used in this Work}\label{sec:metric}
\paragraph{\textbf{Feature Importance.}} In this study, we employ Normalized Mutual Information (NMI) to assess the importance of various features, as mutual information can capture dependencies between features and targets. We implement NMI using the sklearn Python package. Specifically, for classification tasks, we utilize the "feature\_selection.mutual\_info\_classif" function, and for regression tasks, we utilize the "feature\_selection.mutual\_info\_regression" function.
\paragraph{\textbf{Average Normalized AUC across Datasets.}} To aggregate the model performances across datasets, we calculate the average normalized scores~\citep{wistuba2015learning} for AUC to comprehensively evaluate the model performances. Specifically, we first normalize the scores among the compared models for given datasets, and then average them across datasets. Formally, among $D$ involved datasets, the average normalized score $s_m$ for the model $m$ is computed by:
\begin{equation}
    s'_{m, d} = \frac{s_{m, d} - \text{min}_{i \in M_0}(s_{i, d})}{\text{max}_{i \in M_0}(s_{i, d}) - \text{min}_{i \in M_0}(s_{i, d})} , s_m = \frac{\sum^D_{d=1} s'_{m, d}}{D}
\end{equation}
where $M_0$ encompasses all the models compared. The $s_m$ denotes $\text{AUC}_m$. We only use binary classification datasets in Fig.~\ref{fig:rotation_boxplot} since the average normalized AUC, ACC, and nRMSE should be separately computed.

\paragraph{\textbf{Performance Rank.}} We performed 5 runs with different random seeds and calculated the average results for each dataset. Additionally, we computed the overall rank across datasets for comparison. Average rank is given to tied values.

\section{Detailed Description of Datasets}\label{app:datasets}
The details of the 96 used small-scale tabular datasets are summarized in Table~\ref{tab:smalldata1} and Table~\ref{tab:smalldata2}. The details of the 21 large-scale datasets are summarized in Table~\ref{tab:largedatainfo}. We use the same train-valid-test split for all the approaches.
\begin{table*}[ht]
\centering
\caption{The details of the 96 small-scale tabular datasets used. ``\#. Num'' and ``\#. Cat'' denote the numbers of numerical and categorical features, respectively. ``\#.~Sample'' presents the size of a dataset.}\label{tab:smalldata1}
\begin{tabular}{lccccc}
\hline
\multicolumn{1}{c}{Dataset}                                             & \#. Sample & \#. Feature & \#. Num & \#.Cat & Task Type  \\ \hline
Analytics Vidhya Loan Prediction                                 & 614        & 11          & 5       & 6      & classification   \\
Audit Data                                                       & 776        & 24          & 21      & 3      & classification   \\
Automobiles                                                      & 201        & 25          & 13      & 12     & classification   \\
Bigg Boss India                                                  & 567        & 21          & 6       & 15     & classification   \\
Breast Cancer Dataset                                            & 569        & 30          & 30      & 0      & classification   \\
Campus Recruitment                                               & 215        & 13          & 6       & 7      & classification   \\
chronic kidney disease                                           & 400        & 13          & 9       & 4      & classification   \\
House Price & 506        & 17          & 14      & 3      & classification   \\
Compositions of Glass                                            & 214        & 9           & 9       & 0      & classification   \\
Credit Card Approval                                             & 590        & 15          & 6       & 9      & classification   \\
Customer Classification                                          & 1000       & 11          & 5       & 6      & classification   \\
Development Index                                                & 225        & 6           & 6       & 0      & classification   \\
fitbit dataset                                                   & 457        & 13          & 12      & 1      & classification   \\
Horse Colic Dataset                            & 299        & 27          & 9       & 18     & classification   \\
Penguins Classified                                              & 344        & 6           & 4       & 2      & classification   \\
Pima-Indians\_Diabetes                      & 768        & 8           & 8       & 0      & classification   \\
Real Estate DataSet                                              & 511        & 13          & 11      & 2      & classification   \\
Startup Success Prediction                                       & 923        & 45          & 9       & 36     & classification   \\
Store Data Performance                                           & 135        & 16          & 7       & 9      & classification   \\
The Estonia Disaster Passenger List                              & 989        & 6           & 1       & 5      & classification   \\
AAPL\_stock\_price\_2021\_2022                                   & 346        & 5           & 5       & 0      & regression \\
AAPL\_stock\_price\_2021\_2022\_1                                & 347        & 5           & 5       & 0      & regression \\
AAPL\_stock\_price\_2021\_2022\_2                                & 348        & 5           & 5       & 0      & regression \\
analcatdata\_creditscore                                         & 100        & 6           & 3       & 3      & classification   \\
analcatdata\_homerun                                             & 162        & 26          & 12      & 14     & regression \\
analcatdata\_lawsuit                                             & 264        & 4           & 3       & 1      & classification   \\
analcatdata\_vineyard                                            & 468        & 3           & 1       & 2      & regression \\
auto\_price                                                      & 159        & 15          & 13      & 2      & regression \\
autoPrice                                                        & 159        & 15          & 14      & 1      & regression \\
bodyfat                                                          & 252        & 14          & 14      & 0      & regression \\
boston                                                           & 506        & 13          & 11      & 2      & regression \\
boston\_corrected                                                & 506        & 19          & 15      & 4      & regression \\
Boston-house-price-data                                          & 506        & 13          & 11      & 2      & regression \\
cholesterol                                                      & 303        & 13          & 7       & 6      & regression \\
cleveland                                                        & 303        & 13          & 7       & 6      & regression \\
cloud                                                            & 108        & 5           & 3       & 2      & regression \\
cps\_85\_wages                                                   & 534        & 10          & 3       & 7      & regression \\
cpu                                                              & 209        & 7           & 5       & 2      & regression \\
DEE                                                              & 365        & 6           & 6       & 0      & regression \\
Diabetes-Data-Set                                                & 768        & 8           & 8       & 0      & classification   \\
DiabeticMellitus                                                 & 281        & 97          & 6       & 91     & classification   \\
disclosure\_x\_bias                                              & 662        & 3           & 3       & 0      & regression \\
disclosure\_x\_noise                                             & 662        & 3           & 3       & 0      & regression \\
disclosure\_x\_tampered                                          & 662        & 3           & 3       & 0      & regression \\
disclosure\_z                                                    & 662        & 3           & 3       & 0      & regression \\
echoMonths                                                       & 130        & 9           & 7       & 2      & regression \\
EgyptianSkulls                                                   & 150        & 4           & 3       & 1      & regression \\
ELE-1                                                            & 495        & 2           & 2       & 0      & regression \\
fishcatch                                                        & 158        & 7           & 5       & 2      & regression \\
Fish-market                                                      & 159        & 6           & 5       & 1      & regression \\ \hline
\end{tabular}
\end{table*}
\begin{table*}[ht]
\centering
\caption{The details of the 96 small-scale tabular datasets used (continued). ``\#. Num'' and ``\#. Cat'' denote the numbers of numerical and categorical features, respectively. ``\#.~Sample'' presents the size of a dataset.}\label{tab:smalldata2}
\begin{tabular}{lccccc}
\hline
\multicolumn{1}{c}{Dataset}                                             & \#. Sample & \#. Feature & \#. Num & \#.Cat & Task Type  \\ \hline
forest\_fires                                                    & 517        & 12          & 8       & 4      & regression \\
Forest-Fire-Area                                                 & 517        & 12          & 8       & 4      & regression \\
fruitfly                                                         & 125        & 4           & 2       & 2      & regression \\
HappinessRank\_2015                                              & 158        & 9           & 8       & 1      & regression \\
Heart\_disease\_classification                                   & 296        & 13          & 7       & 6      & classification   \\
hungarian                                                        & 294        & 13          & 11      & 2      & classification   \\
Indian-Liver-Patient-Patient               & 583        & 11          & 9       & 2      & classification   \\
Intersectional-Bias-Assessment                 & 1000       & 18          & 14      & 4      & classification   \\
liver-disorders                                                  & 345        & 5           & 5       & 0      & regression \\
lowbwt                                                           & 189        & 9           & 2       & 7      & regression \\
lungcancer\_shedden                                              & 442        & 23          & 20      & 3      & regression \\
machine\_cpu                                                     & 209        & 6           & 6       & 0      & regression \\
meta                                                             & 528        & 21          & 16      & 5      & regression \\
nki70.arff                                                       & 144        & 76          & 72      & 4      & classification   \\
no2                                                              & 500        & 7           & 7       & 0      & regression \\
pharynx                                                          & 195        & 10          & 3       & 7      & regression \\
Pima-Indians-Diabetes                                            & 768        & 8           & 8       & 0      & classification   \\
pm10                                                             & 500        & 7           & 7       & 0      & regression \\
Pokmon-Legendary-Data                                            & 801        & 12          & 9       & 3      & classification   \\
Reading\_Hydro                                                   & 1000       & 26          & 11      & 15     & regression \\
residential\_building                                            & 372        & 108         & 100     & 8      & regression \\
rmftsa\_ladata                                                   & 508        & 10          & 10      & 0      & regression \\
strikes                                                          & 625        & 6           & 6       & 0      & regression \\
student-grade-pass-or-fail-prediction                            & 395        & 29          & 4       & 25     & classification   \\
Swiss-banknote-conterfeit-detection                              & 200        & 6           & 6       & 0      & classification   \\
The-Estonia-Disaster-Passenger-List                              & 989        & 6           & 1       & 5      & classification   \\
The-Office-Dataset                                               & 188        & 10          & 2       & 8      & regression \\
tokyo1                                                           & 959        & 44          & 42      & 2      & classification   \\
visualizing\_environmental                                       & 111        & 3           & 3       & 0      & regression \\
weather\_ankara                                                  & 321        & 9           & 9       & 0      & regression \\
wisconsin                                                        & 194        & 32          & 32      & 0      & regression \\
yacht\_hydrodynamics                                             & 308        & 6           & 6       & 0      & regression \\
Absenteeism at work                                                 & 740        & 20          & 7       & 13     & classification   \\
Audit Data                                                          & 776        & 24          & 21      & 3      & classification   \\
Breast Cancer Coimbra                                               & 116        & 9           & 9       & 0      & classification   \\
Cervical cancer (Risk Factors)                                      & 858        & 30          & 25      & 5      & classification   \\
Climate Model Simulation Crashes                                    & 540        & 19          & 18      & 1      & classification   \\
Early stage diabetes risk prediction                        & 520        & 16          & 1       & 15     & classification   \\
extention of Z-Alizadeh sani dataset                                & 303        & 57          & 20      & 37     & classification   \\
HCV data                                                            & 615        & 12          & 11      & 1      & classification   \\
Heart failure clinical records                                      & 299        & 12          & 7       & 5      & classification   \\
Parkinson Dataset                 & 240        & 46          & 44      & 2      & classification   \\
QSAR Bioconcentration classes                               & 779        & 11          & 7       & 4      & classification   \\
Quality Assessment of DC                          & 97         & 62          & 62      & 0      & classification   \\
User Knowledge Modeling                                             & 258        & 5           & 5       & 0      & classification   \\
Z-Alizadeh Sani                                                     & 303        & 54          & 20      & 34     & classification   \\ \hline
\end{tabular}
\end{table*}
\begin{table*}[t]
\setlength\tabcolsep{3pt}
\centering
\caption{The details of 21 large-scale datasets used. ``\#. Num'' and ``\#. Cat'' denote the numbers of numerical and categorical features, respectively. ``\#.~Sample'' presents the size of a dataset.}\label{tab:largedatainfo}
\resizebox{\linewidth}{!}{%
\begin{tabular}{lccccccp{50mm}}
\toprule
\multicolumn{1}{c}{Dataset}     & Abbr. & \multicolumn{1}{c}{Task Type} & \multicolumn{1}{c}{\#. Features} & \multicolumn{1}{c}{\#. Num} & \multicolumn{1}{c}{\#. Cat} & \multicolumn{1}{c}{\#. Sample} & \multicolumn{1}{c}{Link}                                                \\ \midrule
sulfur                          & SU    & regression                                     & 6                              & 6                          & 0                          & 10,081                         & \scriptsize{\url{https://www.openml.org/d/44145}}                                          \\
bank-marketing                  & BA    & classification                               & 7                              & 7                          & 0                          & 10,578                         & \scriptsize{\url{https://www.openml.org/d/44126}}                                          \\
Brazilian\_houses               & BR    & regression                                 & 8                              & 8                          & 0                          & 10,692                         & \scriptsize{\url{https://www.openml.org/d/44141}}                                          \\
eye                             & EY    & multiclass                                 & 26                             & 26                         & 0                          & 10,936                         & \scriptsize{\url{http://www.cis.hut.fi/eyechallenge2005}}                                  \\
MagicTelescope                  & MA    & classification                                & 10                             & 10                         & 0                          & 13,376                         & \scriptsize{\url{https://www.openml.org/d/44125}}                                          \\
Ailerons                        & AI    & regression                            & 33                             & 33                         & 0                          & 13,750                         & \scriptsize{\url{https://www.openml.org/d/44137}}                                          \\
pol                             & PO    & regression                          & 26                             & 26                         & 0                          & 15,000                         & \scriptsize{\url{https://www.openml.org/d/722}}                                            \\
binarized-pol                   & BP    & classification                                   & 48                             & 48                         & 0                          & 15,000                         & \scriptsize{\url{https://www.openml.org/d/722}}                                            \\
credit                          & CR    & classification                              & 10                             & 10                         & 0                          & 16,714                        & \scriptsize{\url{https://www.openml.org/d/44089}}                                          \\
california                      & CA    & regression                           & 8                              & 8                          & 0                          & 20,640                        & \scriptsize{\url{https://www.dcc.fc.up.pt/~ltorgo/Regression/cal\_housing.html}}      \\
house\_sales                    & HS    & regression                         & 15                             & 15                         & 0                          & 21,613                        & \scriptsize{\url{https://www.openml.org/d/44144}}                                          \\
house                           & HO    & regression                                   & 16                             & 16                         & 0                          & 22,784                        & \scriptsize{\url{https://www.openml.org/d/574}}                                            \\
diamonds                        & DI    & regression                                 & 6                              & 6                          & 0                          & 53,940                        & \scriptsize{\url{https://www.openml.org/d/44140}}                                          \\
helena                          & HE    & multiclass                            & 27                             & 27                         & 0                          & 65,196                        & \scriptsize{\url{https://www.openml.org/d/41169}}                                          \\
jannis                          & JA    & multiclass                               & 54                             & 54                         & 0                          & 83,733                        & \scriptsize{\url{https://www.openml.org/d/41168}}                                          \\
higgs-small                     & HI    & classification                                     & 28                             & 28                         & 0                          & 98,049                        & \scriptsize{\url{https://www.openml.org/d/23512}}                                          \\
road-safety                     & RO    & classification                                  & 32                             & 29                         & 3                          & 111,762                        & \scriptsize{\url{https://www.openml.org/d/44161}}                                          \\
medical\-charges                & ME    & regression                                 & 3                              & 3                          & 0                          & 163,065                       & \scriptsize{\url{https://www.openml.org/d/44146}}                                          \\
SGEMM\_GPU\_kernel\_performance & SG    & regression                         & 9                              & 3                          & 6                          & 241,600                       & \scriptsize{\url{https://www.openml.org/d/44069}}                                          \\
covtype                         & CO    & multiclass                                   & 54                             & 54                         & 0                          & 581,012                       & \scriptsize{\url{https://www.openml.org/d/1596}}                                           \\
nyc-taxi-green-dec-2016         & NY    & regression                            & 9                              & 9                          & 0                          & 581,835                       & \scriptsize{\url{https://www.openml.org/d/44143}}  \\ 
\bottomrule                                       
\end{tabular}
}
\end{table*}
\section{Detailed Results on Small and Large Datasets}\label{app:performances}
We present the average results (five runs averaged) of all the models for each dataset. The results for the 96 small-scale datasets can be found in Table~\ref{tab:small_rawperformance}, and the performance on the 21 large-scale datasets is provided in Table~\ref{tab:large_rawperformances}.
\begin{table*}[ht]
\centering
\setlength{\tabcolsep}{2pt}
\caption{Performance of \model with other state-of-the-art models on 96 public small-scale datasets. O.: \model; O. + F: \model with \featmix; O. + H: \model with \hidmix; XGb: XGboost, Cat: Catboost; FTT: FT-Transformer; TapP: TabPFN; TT: TransTab; XT: XTab. ``(d)'': using default hyperparameters; ``(t)'': hyperparameter fine-tuning is performed. ``(M)'': Mix Tuned version; ``(F)'': Fully Tuned version. TabPFN is designed for classification, we mark ``n/a'' in regression tasks.}\label{tab:small_rawperformance}
\vskip -0.6 em
\resizebox{0.65\textwidth}{!}{
\begin{tabular}{l|cccccccccccccc}
\hline
\multicolumn{1}{c|}{Datasets}                                                   & \multicolumn{1}{c}{O.+F (d)} & \multicolumn{1}{c}{O.+H (d)} & \multicolumn{1}{c}{O. (MT)} & \multicolumn{1}{c}{O.(FT)} & \multicolumn{1}{c}{FTT (t)} & \multicolumn{1}{c}{XGb (t)} & \multicolumn{1}{c}{Cat (t)} & \multicolumn{1}{c}{MLP (t)} & \multicolumn{1}{c}{DCNv2 (t)} & \multicolumn{1}{c}{AutoInt(t)} & \multicolumn{1}{c}{SAINT (t)} & \multicolumn{1}{c}{TT (d)} & \multicolumn{1}{c}{XT(d)} & \multicolumn{1}{c}{TapP(t)} \\ \hline
Analytics   Vidhya Loan Prediction                                 & 0.7449                         & 0.7421                           & 0.7421                        & 0.7421                          & 0.7285                        & 0.7486                        & 0.7045                         & 0.7359                  & 0.7464                    & 0.7350                      & 0.7505                    & 0.7291                       & 0.7240                   & 0.7331                     \\
Audit Data                                                         & 0.9984                         & 0.9905                           & 0.9991                        & 0.9941                          & 0.9993                        & 0.9995                        & 1.0000                         & 0.9936                  & 0.9990                    & 0.9983                      & 0.9995                    & 0.9983                       & 0.9822                   & 0.9998                     \\
\hline
\end{tabular}}
\end{table*}

\begin{table*}[ht]
\centering
\setlength{\tabcolsep}{2pt}
\caption{Performance of \model with other state-of-the-art models on 21 public large-scale datasets. Excel: \model; Excel + F: \model with \featmix; Excel + H: \model with \hidmix; XGb: XGboost, Cat: Catboost; ``(d)'': using default hyperparameters; ``(t)'': hyperparameter fine-tuning is performed. ``(M)'': Mix Tuned; ``(F)'': Fully Tuned.}\label{tab:large_rawperformances}
\resizebox{\textwidth}{!}{
\begin{tabular}{c|ccccc|ccccc}
\hline
Datasets & XGb (d)    & Cat (d)    & FTT (d)    & Excel + F (d) & Excel + H (d) & XGb (t)    & Cat (t)    & FTT (t)    & Excel (M) & Excel (F) \\ \hline
SU       & -0.02025   & -0.01994   & -0.01825   & -0.01840                                        & -0.01740                                       & -0.01770   & -0.02200   & -0.01920   & -0.01730                          & -0.01610                            \\
BA       & 80.25      & 89.20      & 88.26      & 89.00                                           & 88.65                                          & 88.97      & 89.16      & 88.64      & 89.21                             & 89.16                               \\
BR       & -0.07667   & -0.07655   & -0.07390   & -0.11230                                        & -0.06960                                       & -0.07690   & -0.09310   & -0.07940   & -0.06270                          & -0.06410                            \\
EY       & 69.97      & 69.85      & 71.06      & 71.44                                           & 72.09                                          & 72.88      & 72.41      & 71.73      & 74.14                             & 78.94                               \\
MA       & 86.21      & 93.83      & 93.66      & 93.38                                           & 93.66                                          & 93.69      & 93.66      & 93.69      & 94.04                             & 94.11                               \\
AI       & -0.0001669 & -0.0001652 & -0.0001637 & -0.0001689                                      & -0.0001627                                     & -0.0001605 & -0.0001616 & -0.0001641 & -0.0001615                        & -0.0001612                          \\
\hline
\end{tabular}}
\end{table*}
\end{document}